\documentclass[letterpaper]{article} 
\usepackage{aaai2026}  
\usepackage{times}  
\usepackage{helvet}  
\usepackage{courier}  
\usepackage[hyphens]{url}  
\usepackage{graphicx} 
\usepackage{natbib}  
\usepackage{caption} 
\usepackage{amsmath,amssymb,amsfonts,booktabs,multirow,microtype}
\usepackage{tikz}
\usetikzlibrary{arrows.meta,positioning,shapes.geometric}

\title{Embracing Ambiguity: Bayesian Nonparametrics and Stakeholder Participation for Ambiguity‑Aware Safety Evaluation}

\author{Yanan Long}
\affiliations{
    StickFlux Labs \\
    ylong@uchicago.edu
}

\begin{document}
\maketitle

\begin{abstract}
Evaluations of generative AI models often collapse nuanced behaviour into a single number computed for a single decoding configuration. Such point estimates obscure tail risks, demographic disparities, and the existence of multiple near‑optimal operating points. We propose a unified framework that embraces multiplicity by modelling the distribution of harmful behaviour across the entire space of decoding knobs and prompts, quantifying risk through tail‑focused metrics, and integrating stakeholder preferences. Our technical contributions are threefold: (i) we formalise \emph{decoding Rashomon sets}, regions of knob space whose risk is near‑optimal under given criteria—and measure their size and disagreement; (ii) we develop a dependent Dirichlet process (DDP) mixture with stakeholder‑conditioned stick‑breaking weights to learn multi‑modal harm surfaces; and (iii) we introduce an active sampling pipeline that uses Bayesian deep learning surrogates to explore knob space efficiently. Our approach bridges multiplicity theory, Bayesian nonparametrics, and stakeholder-aligned sensitivity analysis, paving the way for trustworthy deployment of generative models.

\end{abstract}

\section{Introduction}

Large generative models now participate in education, healthcare, policy support and creative industries, yet they often produce outputs that are toxic, biased or factually incorrect \citep{Bender2021Dangers,Weidinger2022Taxonomy}.  Evaluation protocols typically fix a handful of decoding settings—temperature, top‑$p$, repetition penalty or model family indicators—generate a few samples per prompt, compute average toxicity scores, and declare success if the average risk is below a threshold.  This practice tacitly assumes that (i) there is a unique \emph{best} operating point and (ii) the mean is the only relevant risk functional.  Both assumptions are problematic.  Real tasks exhibit \emph{predictive multiplicity}: many decoding configurations yield similar average metrics but differ dramatically in tail behaviour and fairness across demographic slices \citep{Breiman2001TwoCultures}.  Policies that ignore such multiplicity can inadvertently select an operating point with unacceptable tail risk or bias.  Recent work in interpretable machine learning has formalised multiplicity through \emph{Rashomon sets}: the collection of models with near‑optimal performance \citep{rudin2024position}.  Motivated by this perspective, we argue that safety evaluation of generative models should recognise and report multiplicity rather than suppress it.

Our goal is thus not to return a single toxicity score but to infer a \emph{posterior distribution over risk surfaces} across knob space and prompts, then compute tail‑sensitive and stakeholder‑specific risk summaries.  This requires three innovations.  First, we formalise \emph{decoding Rashomon sets} as subsets of knob space where a risk functional (mean, conditional value at risk, or disparity gap) is within $\varepsilon$ of its optimum; we propose measures of their size and how they differ across stakeholders.  Second, we develop a dependent Dirichlet process (DDP) mixture model to capture multi‑modal harm distributions that vary with both decoding settings and prompt features.  Unlike parametric logistic surfaces or standard Gaussian Processes which may smooth over distinct failure modes, the DDP yields a flexible family of conditional distributions and naturally expresses multiplicity via its posterior.  Third, we design an active sampling and stakeholder simulation pipeline that calibrates automated toxicity detectors via synthetic human ratings, uses Bayesian deep neural networks to guide exploration of knob space, and employs conformal prediction wrappers for finite‑sample coverage.

Our framework treats stakeholder preferences as first‑class variables that define prompt and knob distributions and evaluation criteria.  While direct human participation is the gold standard, recruiting diverse raters is often expensive and ethically sensitive. To address this, we present a simulation protocol grounded in public datasets (RealToxicityPrompts, Civil Comments, BOLD, SBIC). This protocol serves not as a replacement for human judgment, but as a rigorous \emph{sensitivity analysis} to identify contentious design regions before engaging real communities. These agents draw prompts from topic mixtures and rate outputs with demographic‑specific noise, bias and severity derived from empirical annotation distributions.  We calibrate automated judges using these labels to obtain human‑aligned harm probabilities.  We then fit the DDP mixture to learn multi‑modal harm surfaces and compute risk measures such as conditional value at risk (CVaR), worst‑slice gap, safe‑set volume, and disagreement index.  By integrating stakeholder policies over knob space, we report stakeholder‑specific risk and identify knob regions that are simultaneously safe and high‑utility.

Recent advances in Bayesian evaluation of LLMs motivate our approach.  Bayesian evaluation of LLM behaviour models stochastic generation as a Beta–Binomial process and uses sequential sampling to reduce evaluation cost \citep{Longjohn2025Bayesian}.  We extend this to multi‑modal distributions and knob‑continuous risk surfaces.  Additionally, we leverage conformal prediction \citep{Kim2025Adaptive,Wang2025SConU} as a lightweight wrapper to ensure valid uncertainty quantification even under the covariate shift induced by active sampling.  Finally, calls to human‑centered uncertainty quantification emphasise the need to account for aleatoric and distributional uncertainty and to measure utility for actual users \citep{Devic2025HumanCentered}.  Our stakeholder simulation responds directly by placing value preferences and usage policies at the core of the risk computation.

The remainder of the paper is organised as follows.  We first provide a formal problem formulation, defining prompts, knob space, harm variables and stakeholders.  We then lay out our four‑stage pipeline: space‑filling design and active sampling, stakeholder prior elicitation and policy specification, simulated judging and calibration, and Bayesian inference with risk reporting.  We present the full DDP generative model, describe Bayesian deep learning surrogates for active sampling and logistic calibration, and derive conformal wrappers for finite‑sample coverage.  We discuss simulation details and relate our work to multiplicity, fairness and uncertainty literature.  We conclude with experiments on synthetic and real LLM evaluations and outline future directions.

\section{Formal Problem Statement and Risk Functionals}

\subsection{Generative model and decoding knobs}

Let $x \in \mathcal{X} \subset \mathbb{R}^d$ denote a vector of decoding settings, such as temperature, nucleus sampling probability, repetition penalty, model family indicators, or refusal filter toggles.  The generative model $\pi$ defines a stochastic mapping from prompts $p \in \mathcal{P}$ and knobs $x$ to outputs $y \sim \pi(\cdot \mid p, x)$.  Because decoding is random, each $(p,x)$ pair induces a full distribution over outputs.  We assume a finite set of prompts $\{p_i\}_{i=1}^{I}$ drawn either from a stakeholder‑defined distribution $\mathbb{P}(p \mid s)$ where $ s \in \mathcal{S} $ is the set of stakeholders, or from a simulation process described later.

We are interested in harmful behaviour captured by a categorical (with binary as a special case), ordinal or continuous harm score $H(y)$.  For example, $H$ could be a toxicity probability, a measure of privacy leakage, or the indicator of a jailbreak.  In practice $H$ is not directly observable but is estimated via a judge—either an automated classifier or a human rater.  We denote the calibrated harm score by $\tilde{h}(y) \in [0,1]$ (or the simplex $\Delta^{C-1}$ for categorical) and treat it as an estimate of $\mathbb{P}(\text{harm}\mid y)$.

\subsection{Stakeholders}

A stakeholder $s$ is characterised by three ingredients: (i) a distribution over prompts $p(p \mid s)$ reflecting their domain of interest; (ii) a distribution over decoding knobs $p(x \mid s)$ reflecting typical usage or policy constraints; and (iii) a risk threshold or preference functional.  Stakeholders may represent content moderators, product managers, developers, or different demographic user groups.  When real participants are unavailable, we simulate stakeholders by sampling their prompt topics, knob policies and sensitivities from priors anchored to public datasets and demographic research (see Section~\ref{sec:simulation}).

\subsection{Risk functionals and multiplicity metrics}

Let $Z_{p,x}$ denote the predictive harm random variable at prompt $p$ and knob setting $x$.  The \emph{mean harm surface} is $\mu(p,x) = \mathbb{E}[Z_{p,x}]$.  To capture tail risk, we define the conditional value at risk (CVaR) at level $\alpha \in (0,1)$,
\begin{align}
    \mathrm{CVaR}_{\alpha}(p,x) = \frac{1}{1-\alpha}\,\mathbb{E} \bigl[Z_{p,x}\,\big|\,Z_{p,x} \ge F_{p,x}^{-1}(\alpha)\bigr],
    \label{eq:cvar}
\end{align}
where $F_{p,x}$ is the distribution function of $Z_{p,x}$ \citep{Rockafellar2000}.  CVaR measures expected harm in the worst $1-\alpha$ fraction of outcomes.  For fairness, let $\mathcal{G}$ be a set of demographic slices (e.g., identity groups).  The harm rate for group $g$ at $x$ is $\mu_g(x) = \mathbb{E}[Z_{p,x}\,\mathbb{1}\{p \in g\}]/\mathbb{P}(p \in g)$.  The \emph{worst‑slice gap} is
\begin{align}
    \mathrm{Gap}(x) = \max_{g \in \mathcal{G}} \mu_g(x) - \min_{g \in \mathcal{G}} \mu_g(x).
    \label{eq:gap}
\end{align}
Finally, for a stakeholder $s$ with knob policy $p(x \mid s)$, the expected harm is
\begin{align}
    R_s = \int_{\mathcal{X}} \mu(p(s),x)\,p(x \mid s)\,\mathrm{d}x,
    \label{eq:stakeholder-risk}
\end{align}
where $p(s)$ indicates that the prompt distribution depends on $s$.  These functionals define the targets we wish to estimate.

\paragraph{Decoding Rashomon sets and disagreement metrics.}  Fix a risk functional $R(x)$ (e.g., $\mathrm{CVaR}_{\alpha}(x)$).  The $\varepsilon$‑Rashomon set is
\begin{align}
    \mathcal{R}_{\varepsilon} = \{x \in \mathcal{X}: R(x) \le R^{\star} + \varepsilon\},
    \label{eq:rashomon-set}
\end{align}
where $R^{\star} = \min_{x} R(x)$ or any baseline.  We measure multiplicity via the \emph{safe volume}
\begin{align}
    \mathrm{Vol}_{\varepsilon} = \mathrm{vol}\bigl(\mathcal{R}_{\varepsilon}\bigr)/\mathrm{vol}(\mathcal{X}),
    \label{eq:volume}
\end{align}
where $\mathrm{vol}$ denotes Lebesgue measure, and the \emph{disagreement index}
\begin{align}
    \mathrm{Disagree}(x) = \operatorname{Var}_{s} \bigl(R_s(x)\bigr),
    \label{eq:disagreement}
\end{align}
 which quantifies how stakeholder risk differs at $x$.  Posterior distributions over $\mathcal{R}_{\varepsilon}$ and $\mathrm{Vol}_{\varepsilon}$ capture epistemic uncertainty about multiplicity.

These risk functionals emphasise distinct aspects of harmful behaviour.  The mean $\mu(p,x)$ measures the expected harm across stochastic generations, integrating over aleatoric uncertainty; however, for rare but catastrophic failures the mean can be small even when the tail is unacceptable.  CVaR$_{\alpha}$ therefore focuses on the worst $1-\alpha$ fraction of outcomes and is widely used in finance and risk management as a coherent risk measure \citep{Rockafellar2000}.  In our context, $\mathrm{CVaR}_{0.95}(p,x)$ quantifies expected toxicity among the top $5\%$ most harmful completions for a prompt/knob pair.  The worst‑slice gap~\eqref{eq:gap} quantifies fairness by comparing the highest and lowest mean harm rates across demographic slices $\mathcal{G}$; a large gap indicates that some identity groups are disproportionately harmed.  Stakeholder risk $R_s$ aggregates the mean harm over a stakeholder's knob policy and prompt distribution, weighting $\mu(p,x)$ by how likely the stakeholder is to encounter each configuration.

Besides these metrics, we also consider exceedance probabilities and quantiles.  Let $q_{\beta}(p,x)$ denote the $\beta$‑quantile of $Z_{p,x}$ (the inverse of $F_{p,x}$); then the exceedance probability $\mathbb{P}(Z_{p,x} \geq \tau)$ and quantile $q_{\beta}(p,x)$ can be estimated from posterior draws of $Z$.  These quantities support threshold‑based safety policies (e.g., “at most $5\%$ of completions may exceed a toxicity threshold”) and can be used to define Rashomon sets based on exceedance constraints.

When summarising risk across prompts, we integrate over the stakeholder's prompt distribution: $\bar{\mu}(x) = \mathbb{E}_{p \sim p(p\mid s)}[\mu(p,x)]$ and analogously $\overline{\mathrm{CVaR}}_{\alpha}(x)$.  These collapsed surfaces drive the stakeholder risk~\eqref{eq:stakeholder-risk} and are crucial when stakeholders have different prompt preferences.

The safe volume~\eqref{eq:volume} captures how much of knob space satisfies a safety criterion.  A large safe volume indicates that many settings are essentially equivalent in risk (high multiplicity), while a tiny safe volume suggests a narrow “sweet spot.”  The disagreement index~\eqref{eq:disagreement} measures how stakeholder utilities diverge; high variance implies that different groups perceive the same knob setting very differently, signalling value multiplicity.  In Section~\ref{sec:simulation} we describe simulation protocols that induce such divergences.

\section{Methodology}

Our evaluation pipeline comprises four stages (Figure~\ref{fig:pipeline}).

\begin{figure}[t]
    \centering
    \begin{tikzpicture}[
        node distance=0.5cm,
        stage/.style={draw, rounded corners=3pt, minimum width=0.85\linewidth, minimum height=1.2cm, align=center, font=\small, fill=black!5},
        arrow/.style={-{Stealth[length=2.5mm]}, thick},
        label/.style={font=\scriptsize, text=black!70, fill=white, inner sep=1pt}
    ]
    \node[stage] (s1) {
        \textbf{Stage 1: Design \& Sampling}\\[2pt]
        \scriptsize $x_n \sim \text{Sobol}, \quad \hat{\mu}_t, \hat{v}_t \to a_t(x)$
    };
    
    \node[stage, below=of s1] (s2) {
        \textbf{Stage 2: Stakeholder Priors}\\[2pt]
        \scriptsize Policies $p(x\mid s), \quad \text{Topics } \theta_g$
    };
    
    \node[stage, below=of s2] (s3) {
        \textbf{Stage 3: Judging \& Calibration}\\[2pt]
        \scriptsize Categorical/Ordinal/Continuous, $\tilde{h} = f(J;\beta)$
    };
    
    \node[stage, below=of s3] (s4) {
        \textbf{Stage 4: DDP Inference}\\[2pt]
        \scriptsize $z \sim \pi_k, \quad \tilde{h} \mid z \sim F_k$
    };
    
    \node[stage, below=of s4, fill=black!10] (out) {
        \textbf{Risk Report}\\[2pt]
        \scriptsize $\mathcal{R}_\varepsilon, \mathrm{Vol}_\varepsilon, \mathrm{CVaR}_\alpha$
    };
    
    \draw[arrow] (s1) -- node[label] {Outputs $y_{ijr}$} (s2);
    \draw[arrow] (s2) -- node[label] {Simulated Raters} (s3);
    \draw[arrow] (s3) -- node[label] {Harm Scores $\tilde{h}$} (s4);
    \draw[arrow] (s4) -- (out);
    
    \node[left=0.2cm of s1, font=\scriptsize, align=right, text=black!60] (in1) {LLM $\pi$};
    \draw[arrow, black!60] (in1) -- (s1);
    
    \draw[arrow, dashed, black!60] (s1.east) -- ++(0.2,0) |- node[pos=0.25, right, font=\tiny, text=black!50] {Active Loop} (s1.north east);
    
    \end{tikzpicture}
    \caption{Vertical evaluation pipeline. Stage 1 generates outputs using active sampling. Stage 2 defines stakeholder priors and rater profiles. Stage 3 simulates judging and calibrates scores. Stage 4 infers risk surfaces and Rashomon sets.}
    \label{fig:pipeline}
\end{figure}

\subsection{Stage 1: Design, active sampling and prompts}

Given a bounded knob domain $\mathcal{X} \subset [0,1]^d$ and evaluation budget $B$, we must choose a finite set of configurations $\{x_j\}$ and replicate counts $R_j$ that efficiently characterise the harm surface across $\mathcal{X}$.  We begin with a \emph{space‑filling design} based on low‑discrepancy sequences (Sobol or Halton).  These sequences generate $N$ initial points $x_1,\dots,x_N$ that minimise the discrepancy between the empirical distribution of $\{x_j\}$ and the uniform measure on $\mathcal{X}$.  The number of points $N$ and replicates per point $R_j$ satisfy $B \approx \sum_j R_j I$, where $I$ is the number of prompt samples; typical practice uses a fixed replicate count $R_j=R$ across points to separate aleatoric variability (within‑point variance) from epistemic uncertainty (between‑point variability).  For each $(p_i, x_j)$ pair we draw $R$ independent outputs $y_{ijr}$ from $\pi(\cdot \mid p_i,x_j)$, obtaining an empirical estimate of the harm distribution at that configuration.

\paragraph{Adaptive sampling and Bayesian surrogate.}  Once initial points have been evaluated, we refine the design using a Bayesian neural network (BNN) to approximate the latent harm surface.  Because the generative model is treated as a black box, we cannot differentiate with respect to knobs. Instead, we train a neural network on observed harm labels to predict the mean harm and quantify epistemic uncertainty at unobserved $(p,x)$.  Denote by $\xi(p,x)$ the concatenation of a prompt embedding and a standardised knob vector.  We posit a BNN $f_W(\xi)$ with weight prior $p(W)$ and define a latent score and harm probability by
\begin{align}
    \eta(p,x) &= f_W\bigl(\xi(p,x)\bigr), \quad W \sim p(W),\label{eq:bdn-def} \\
    \mu(p,x) &= \sigma\bigl(\eta(p,x)\bigr),\label{eq:bdn-mu}
\end{align}
with $\sigma$ the softmax function for categorical harm (reducing to logistic for binary) or the identity for continuous scores.  We perform approximate Bayesian inference (e.g., MC dropout or deep ensembles) to obtain weight samples $\{W^{(m)}\}_{m=1}^M$ and compute the predictive mean and variance
\begin{align}
    \hat{\mu}_t(p,x) &= \frac{1}{M}\sum_{m=1}^M \sigma\bigl(f_{W^{(m)}}(\xi(p,x))\bigr),\label{eq:bdn-mean}\\
    \hat{v}^{\mathrm{epi}}_t(p,x) &= \frac{1}{M-1}\sum_{m=1}^M \Bigl(\sigma\bigl(f_{W^{(m)}}(\xi(p,x))\bigr) - \hat{\mu}_t(p,x)\Bigr)^2.\label{eq:bdn-var}
\end{align}

For Bernoulli (binary) or Categorical labels, the total predictive variance combines epistemic uncertainty and intrinsic variability:
\begin{align}
    \hat{v}_t(p,x) \approx \hat{v}^{\mathrm{epi}}_t(p,x) + \operatorname{diag}(\hat{\mu}_t(p,x)) - \hat{\mu}_t(p,x)\hat{\mu}_t(p,x)^\top,
    \label{eq:bdn-total-var}
\end{align}
(or the scalar variance $\hat{\mu}(1-\hat{\mu})$ in the binary case).  If labels are continuous, an additional network head predicts heteroscedastic noise and its expectation is added to $\hat{v}^{\mathrm{epi}}_t$ to form the total predictive variance.  These surrogate statistics play two roles: they serve as learned features for the BNP mixture weights (Stage 4) and they drive adaptive sampling over knobs.

To explore $\mathcal{X}$, we draw a large candidate set from a low‑discrepancy Sobol sequence $\{u_n\}$ on $[0,1]^d$ and map each point to a knob via $x=T(u)$, where $T$ applies the inverse CDF of the stakeholder’s knob policy $p(x\mid s)$ (see Stage 2).  For each candidate $x$ we sample a batch of prompts $\{p^{(\ell)}\}$ from $p(p\mid s)$ and compute a threshold‑oriented acquisition score for each pair:
\begin{align}
    a_t\bigl(p^{(\ell)},x\bigr) = \frac{\sqrt{\mathrm{tr}\,\hat{v}_t(p^{(\ell)},x)}}{\bigl\|\hat{\mu}_t(p^{(\ell)},x) - \tau\bigr\|_1 + \epsilon},
    \label{eq:stage1-straddle}
\end{align}
where $\tau$ is a risk threshold vector (or scalar for binary/continuous), $\mathrm{tr}$ denotes the trace (reducing to scalar variance in the binary case), and $\epsilon>0$ prevents division by zero.  We aggregate acquisition over prompts by averaging:
\begin{align}
    \bar{a}_t(x) = \frac{1}{L}\sum_{\ell=1}^L a_t\bigl(p^{(\ell)},x\bigr).
    \label{eq:avg-acquisition}
\end{align}
Configurations with high $\bar{a}_t$ either exhibit high epistemic uncertainty or lie close to the risk threshold.  We select the top candidate(s) according to $\bar{a}_t$ for expensive evaluation with the LLM and update the surrogate with the new observations.  Replicate counts $R_j$ are chosen using a variance decomposition: the variance of the sample mean $\hat{\theta}_j$ decomposes into aleatoric and epistemic terms $\mathrm{Var}(\hat{\theta}_j) = \sigma^2_a(x_j)/R_j + \sigma^2_e(x_j)$.  We initialise $R_j$ to a small constant (e.g., $5$) and increase it if the surrogate indicates high epistemic variance or the empirical aleatoric variance is large.

\paragraph{Quasi–Monte Carlo design.}  For high‑dimensional knob spaces, simple random sampling may leave large gaps. Low‑discrepancy sequences such as Sobol, Halton or Niederreiter produce quasi–Monte Carlo points with provably lower star discrepancy than IID samples. Given a $d$‑dimensional space $\mathcal{X} = [0,1]^d$, the $n$th point of a Sobol sequence $x_n$ is constructed by bitwise operations using primitive polynomials. The discrepancy $D_n$ of the first $n$ points satisfies $D_n = O(\frac{(\log n)^d}{n})$, compared with $O(n^{-1/2})$ for uniform random sampling.  We scale and shift $x_n$ to match the bounds of each knob.  To incorporate stakeholder policies $p(x\mid s)$, we perform importance sampling: draw $u_n$ from a Sobol sequence on $[0,1]^d$ and map it to $x_n = F_s^{-1}(u_n)$ using the inverse CDF of $p(x\mid s)$.  This ensures that the initial design reflects the stakeholder’s typical usage while retaining low discrepancy.

\paragraph{Prompt sampling.}  For each stakeholder $s$, we draw prompts $p_i$ from $p(p \mid s)$ the given dataset.  Let $q^{(c)}(p)$ be the empirical distribution of prompts for category $c \in \mathcal{C}$.  A stakeholder from group $g(s)$ with topic mixture $\theta_{g(s)}$ samples a category $c \sim \theta_{g(s)}$ and then draws $p_i \sim q^{(c)}$.  To balance evaluation across categories, we can stratify the initial prompt set so that each category is represented proportionally to its Dirichlet weight and oversample categories with high prior harm rates.  During active sampling, we treat prompts as part of the input to the BDN surrogate and compute acquisition scores jointly over prompt–knob pairs.

Prompts $p_i$ are sampled from the stakeholder's distribution $p(p \mid s)$ (detailed in Stage 2) or, for simulation, from topic mixtures anchored in public corpora.  Each chosen $(p_i,x_j)$ is evaluated $R$ times, producing harm labels $\tilde{h}(y_{ijr})$ for surrogate training and DDP inference.

Beyond guiding sample selection, the Bayesian surrogate offers a useful by-product: a rapid, approximate delineation of safe regions.  By thresholding $\hat{\mu}_t(p,x) + c\sqrt{\hat{v}_t(p,x)} \le \tau$ for an appropriate constant $c$, practitioners can sketch preliminary Rashomon sets in minutes rather than waiting hours for full DDP inference.  This "fast path" is valuable during iterative model development, where quick feedback on which knob regions appear safe accelerates experimentation.  However, the surrogate's unimodal assumptions make it unsuitable for final audits; once exploration concludes, the collected data feed into the DDP mixture (Stage~4), which captures multi-modal failure modes that the surrogate cannot represent.

\subsection{Stage 2: Prior elicitation and stakeholder policies}

Stakeholders contribute two kinds of priors: beliefs about harm rates and policies over knobs and prompts.  Suppose a stakeholder believes that the harm rate $\theta$ for a class of prompts is around $m$ with a credible interval $[l,u]$ at level $1-\alpha$.  We encode this prior with a $\mathrm{Beta}(a,b)$ distribution by solving
\begin{align}
    m &= \frac{a}{a+b},\\
    l &= \mathrm{Beta}^{-1} \bigl(\tfrac{\alpha}{2}; a,b\bigr),\qquad u = \mathrm{Beta}^{-1} \bigl(1-\tfrac{\alpha}{2}; a,b\bigr),
\end{align}
where $\mathrm{Beta}^{-1}$ denotes the quantile function.  Numerical root‑finding yields $(a,b)$; the total pseudo‑count $a+b$ reflects the strength of the prior.  For stakeholders specifying only a mean and a confidence width, we choose $a+b$ via a default rule (e.g., $a+b=2$ or $5$) and solve for $a$ accordingly.

Usage policies over knobs are modelled with product distributions.  For continuous knobs $x_k\in[0,1]$ we assign $\mathrm{Beta}(r_k,s_k)$, where $(r_k,s_k)$ reflect preferences: $r_k > s_k$ implies a preference for high values, while $r_k < s_k$ favours low values.  For discrete knobs we use categorical distributions with Dirichlet priors.  Stakeholders also specify which harm topics they care about.  Let $\mathcal{C}=\{\text{hate},\text{harassment},\text{misinformation},\dots\}$ denote harm categories.  Each demographic group $g$ is associated with Dirichlet parameters $\alpha^{(g)}\in \mathbb{R}_{>0}^{|\mathcal{C}|}$ and draws a topic mixture $\theta_g \sim \mathrm{Dirichlet}(\alpha^{(g)})$.  A stakeholder $s$ from group $g(s)$ samples a harm category $c \sim \theta_{g(s)}$ and then a prompt from a corpus containing category $c$.  This hierarchical model allows different groups to emphasise different categories and ensures that the evaluation reflects stakeholder concerns.

Stakeholder sensitivities and biases in the rating models (Eqs.~\ref{eq:ordinal-model}–\ref{eq:bt-model}) have hierarchical Gaussian priors,
\begin{align}
    a_s \sim \mathcal{N} \bigl(m_a^{(g)}, \sigma_{a}^{2,(g)}\bigr),\quad b_s \sim \mathcal{N} \bigl(m_b^{(g)}, \sigma_{b}^{2,(g)}\bigr),
\end{align}
where the hyperparameters $(m_a^{(g)}, \sigma_a^{2,(g)}, m_b^{(g)}, \sigma_b^{2,(g)})$ capture group‑level tendencies gleaned from annotated datasets \citep{Sap2022AnnotatedAttitudes}.  This hierarchy allows individuals within a group to vary while sharing statistical strength.  When real stakeholders provide prior information (e.g., “I expect harm in $10\%$ of my use cases”), we apply the same Beta or Dirichlet transformation to convert qualitative summaries into statistical priors, ensuring interpretability and transparency.

\subsection{Stage 3: Judging, calibration and simulation}
\label{sec:simulation}

The raw harm score $H(y)$ is estimated via judges.  We use a fast primary judge (e.g., a lightweight classifier or toxicity detector) to provide bulk scores $J(y) \in [0,1]$ for every output.  To account for miscalibration and bias, we collect a small calibration subset with higher-quality labels $r(y)$ from a small LLM judge or human raters.  This split is cost‑effective for 1B–8B evaluations: the primary judge scales across large output volumes, while the small LLM judge only labels a small fraction to correct systematic bias.  Stage 3 supports categorical, ordinal, and continuous harms; the ordinal model below is one option used when ratings are Likert‑scaled.  These ratings may be categorical (e.g., harm types), ordinal (e.g., Likert scales), or pairwise preferences.  For simulated stakeholders, we model a latent harm severity $\zeta(y)$.  In the categorical case (including binary), we sample $r(y) \sim \mathrm{Categorical}(\mathrm{softmax}(W_s \zeta(y) + b_s))$.  For ordinal ratings $r(y) \in \{1,\dots,L\}$, we use:
\begin{align}
    \mathbb{P} \bigl(r(y) \le \ell \mid \zeta(y), s\bigr)
    &= \sigma \left(\tau_{s,\ell} - a_s\,\zeta(y) - b_s\right),
    \label{eq:ordinal-model}
\end{align}
where $\sigma$ is logistic, $a_s$ the stakeholder sensitivity, $b_s$ severity bias and $\tau_{s,\cdot}$ thresholds.  Pairwise preferences are sampled via a Bradley–Terry model,
\begin{align}
    \mathbb{P} \bigl(y_a \succ y_b \mid s\bigr)
    = \sigma \Big(a_s[\zeta(y_a) - \zeta(y_b)] + b_s\Big).
    \label{eq:bt-model}
\end{align}
We calibrate the automated judge using the calibration subset.  Let $J(y)$ denote the raw judge output (e.g., a scalar score in $[0,1]$, a vector of class logits/probabilities, or an ordinal‑logit representation) and let $\tilde{h}(y)$ denote the corresponding calibrated quantity that enters Eq.~\eqref{eq:component-dist}.  For scalar $J(y)$ (continuous or Likert‑style harm scores) we fit a monotone mapping $\tilde{h}(y) = f(J(y);\beta)$ via isotonic regression, a logistic (Platt‑style) map, or an ordinal link.  Concretely, isotonic calibration fits a nondecreasing function $f$ that minimises $\sum_i (r_i - f(J_i))^2$ subject to $f(J_i) \le f(J_j)$ whenever $J_i \le J_j$.  For pairwise labels, we instead fit a Bradley–Terry calibration that maps differences in $J(y)$ to preference probabilities, which are then treated as calibrated observations $\tilde{h}(y)$ for the BNP model.  We place a posterior over $\beta$ (or the corresponding calibration parameters) and propagate uncertainty to the BNP model.  If multiple automated judges exist, we combine them via a Bayesian last‑layer ensemble.

\subsubsection{Harm-modality mapping for calibration and BNP inputs}
We map each harm modality to (i) the judge output type, (ii) the calibration method, and (iii) how calibrated scores enter the BNP mixture:
\begin{itemize}
    \item \textbf{Categorical harm (including binary).} The judge emits class logits or probabilities over harm categories; we calibrate with a multinomial link (temperature scaling or isotonic/Dirichlet calibration) to obtain $\tilde{p}(c \mid y)$.  Categorical labels are produced explicitly by sampling $r(y) \sim \mathrm{Categorical}(\tilde{p}(\cdot \mid y))$ (for simulation) or taking $\arg\max_c \tilde{p}(c \mid y)$ (for deterministic labeling), yielding a discrete label rather than an ordinal or continuous score.  In this case, the calibrated observation is $\tilde{h}(y)=r(y)$, which is used with a $\mathrm{Categorical}(\theta_k)$ component in Eq.~\eqref{eq:component-dist}.
    \item \textbf{Ordinal harm.} The judge outputs an ordinal score or cumulative logits; we calibrate with an ordinal link (e.g., proportional odds) to obtain $\mathbb{P}(r(y)=\ell \mid y)$ over ordered bins.  From this calibrated distribution we form an ordinal label $r(y)$, either by sampling (for simulation) or by taking $\arg\max_\ell \mathbb{P}(r(y)=\ell \mid y)$ (for deterministic labeling), and the BNP mixture then uses $\tilde{h}(y)=r(y)$ as the observation under the $\mathrm{OrdinalLogit}(\tau_k,a_k)$ component in Eq.~\eqref{eq:component-dist}.
    \item \textbf{Continuous harm.} The judge outputs a scalar in $[0,1]$ (or real‑valued), calibrated with isotonic or Platt mapping to $\tilde{h}(y)\in[0,1]$.  These calibrated scores are fed directly into the Beta component of the BNP mixture.
\end{itemize}

\subsection{Stage 4: BNP inference with stakeholder‑conditioned DDP mixtures}

The core of our approach is a flexible generative model for harm conditional on knobs $x$, prompts $p$ and stakeholders $s$.  We posit a latent mixture structure: for each output $y$ we draw a component assignment $z \in \{1,\dots,K\}$, a harm parameter $\theta_k$ determining the distribution of $\tilde{h}(y)$, and mixture weights $\pi_k(x,p,s)$ that depend on the knob, prompt and stakeholder.  Our generative model is:

\begin{align}
    z_{ijr} &\sim \mathrm{Categorical}(\pi_1(x_j, p_i, s),\dots,\pi_K(x_j, p_i, s)),
    \label{eq:comp-assign}\\
    \tilde{h}(y_{ijr}) &\mid \left( z_{ijr} = k \right) \sim
        {\small
        \begin{cases}
            \mathrm{Categorical}(\theta_k), & \text{categorical harm,}\\
            \mathrm{OrdinalLogit}(\tau_k, a_k), & \text{ordinal harm,}\\
            \mathrm{Beta}(\eta_k, \lambda_k), & \text{continuous harm.}
        \end{cases}
        }
    \label{eq:component-dist}
\end{align}
The base measure $H$ for $\theta_k$ can be $\mathrm{Dirichlet}(\alpha)$ (reducing to Beta for binary); for ordinal harm we place priors on the cutpoints $\tau_k$ and slope $a_k$; for Beta‑distributed harm we set $(\eta_k,\lambda_k) \sim \mathrm{Gamma}$.  The mixture weights follow a logistic stick‑breaking construction:
\begin{align}
    v_k(x,p,s) &= \sigma \big(g_k(x,p,s)\big),\quad 
    \pi_k = v_k \prod_{h<k}(1 - v_h).
    \label{eq:stick-break}
\end{align}
The gating function $g_k$ drives the mode selection. We propose a flexible parameterisation:
\begin{align}
    g_k(x,p,s) &= \gamma_k^\top \mathbf{f}(x,p) + \delta_k^\top \rho(s),
    \label{eq:gate-func}
\end{align}
where $\rho(s)$ encodes stakeholder identity and $\mathbf{f}(x,p)$ represents prompt--knob features (e.g., splines or GP kernels).  Note that the DDP operates on the data collected during Stage~1; the Bayesian deep surrogate used for active sampling is separate from the DDP and does not appear in the final inference.

\paragraph{Advanced Modelling.} For scenarios requiring extreme precision in tail risk modelling, we extend this framework with a mean-residual coupling and finite-sample conformal guarantees. These advanced components are detailed in Appendices A and B.

\paragraph{Inference strategy.} We employ a joint variational inference approach to fit the DDP model. We approximate the posterior over mixture parameters $\Theta$ using a factorised family $q(\Theta)$ and maximise the evidence lower bound (ELBO).

\paragraph{Risk reporting.}  For each posterior draw, we compute risk surfaces and multiplicity metrics defined in Section~2.  We sample $x$ from $\mathcal{X}$ and approximate integrals via quadrature or Monte Carlo.  Stakeholder‑specific risks are computed by drawing $x \sim p(\cdot \mid s)$.  We summarise the posterior over safe volumes and disagreement indices, producing credible intervals.  The final \emph{risk report} includes heatmaps of $\mu(p,x)$, $\mathrm{CVaR}_{\alpha}(p,x)$, safe volumes $\mathrm{Vol}_{\varepsilon}$, Rashomon set membership probabilities $\mathbb{P}(x \in \mathcal{R}_{\varepsilon})$, and stakeholder‑specific risk distributions, along with synthetic exemplars of failure modes (mixture component samples).

\section{Stakeholder-Aligned Sensitivity Analysis via Simulation}

We employ simulation to stress-test the safety framework against diverse value systems. This approach allows us to map the "disagreement surface" and identify which knob configurations are robust to conflicting stakeholder priorities. We instantiate synthetic stakeholders as follows.

\paragraph{Demographic groups.}  We define $G$ groups (e.g., \texttt{majority}, \texttt{minority1}, \texttt{minority2}) with proportions $\pi_g$.  Each group $g$ has topic mixture $\theta_g \sim \mathrm{Dirichlet}(\alpha^{(g)})$ over harm categories $\mathcal{C}$ (hate, harassment, misinformation, etc.).  Stakeholder $s$ in group $g$ samples a harm category $c \sim \mathrm{Cat}(\theta_g)$ and draws a prompt $p$ from a category‑specific corpus (e.g., RealToxicityPrompts or BOLD slices) possibly augmented with template transformations.  Topic mixtures produce group‑dependent evaluation distributions.

\paragraph{Knob policies.}  For each group $g$, we sample a Beta distribution over continuous knobs and a categorical distribution over discrete knobs.  For example, $x_{\text{temperature}} \sim \mathrm{Beta}(2,5)$ for conservative groups and $\mathrm{Beta}(5,2)$ for exploratory groups.  Stakeholders thus test the generative model in knob regions reflecting their usage.

\paragraph{Sensitivity and bias.}  Each stakeholder has sensitivity $a_s>0$, baseline bias $b_s$ and noise $\sigma_s$.  We draw $(a_s,b_s)$ from group‑specific priors (e.g., Normal distributions) and $\sigma_s$ from $\mathrm{InverseGamma}$.  To ensure these simulations are not merely arbitrary personas, we ground the priors in real-world data. For instance, we fit the hyperparameters $(m_a^{(g)}, \sigma_a^{2,(g)})$ to the empirical distribution of toxicity ratings found in the Jigsaw Civil Comments dataset, ensuring that the variance and severity thresholds of our synthetic stakeholders match those of real annotators. This treats the simulation as a rigorous sensitivity analysis of the safety evaluation to stakeholder diversity.  We fix thresholds $\tau_{s,\ell}$ to enforce monotone Likert categories.

\paragraph{Calibration sets.}  A calibration dataset of rated outputs is drawn by sampling a subset of $(p,x)$ pairs from Stage~1.  We calibrate the automated judge using ordinal or pairwise models (Section~\ref{sec:simulation}) and propagate uncertainty via sampling.  The calibration set is separate from the evaluation set to avoid double dipping.

\section{Related Work}

Our work builds on four strands of research.

\paragraph{Multiplicity and Rashomon sets.}  The Rashomon effect denotes the existence of many near‑optimal models; predictive multiplicity formalises this as the set of predictors with performance within $\varepsilon$ of the optimum and studies its size and diversity \citep{rudin2024position}.  We extend this concept to generative safety by defining decoding Rashomon sets over knob space and prompt distributions and quantifying their volume and stakeholder disagreement.


\paragraph{Uncertainty quantification for generative AI.}  Conformal prediction has been applied to LLMs to provide distribution‑free coverage for output sets or factuality scores \citep{Angelopoulos2023LLMConformal}.  Adaptive conformal bands partition the predictor space and calibrate locally to tighten intervals \citep{Kim2025Adaptive}, and selective conformal uncertainty tests remove outliers that break exchangeability assumptions \citep{Wang2025SConU}.  We incorporate these methods as wrappers around our Bayesian predictions to guarantee valid uncertainty.

\paragraph{Human‑centered evaluation and fairness.}  Recent critique argues that UQ for LLMs often focuses on epistemic uncertainty, uses benchmarks with low ecological validity, and optimises metrics unrelated to user utility \citep{Devic2025HumanCentered}.  Our stakeholder simulation and multiplicity focus answer this call by integrating aleatoric uncertainty from stochastic decoding, modelling distributional uncertainty across prompts and knobs, and reporting risk metrics aligned with stakeholder values and fairness concerns.

\section{Experiments}

We demonstrate our framework in two settings.

\subsection{Synthetic ground‑truth study}

We construct a synthetic environment with two knobs: temperature $x_1 \in [0,1]$ and top‑$p$ $x_2 \in [0,1]$.  We define a “true” mixture of three harm modes, with mixture weights varying sinusoidally in $x_1$ and $x_2$, and simulate outputs accordingly.  We generate prompts from two topics and calibrate a mis‑calibrated automated judge via simulated stakeholders.  We compare our DDP mixture with (i) a "Lite" baseline using Quantile Regression Forests (QRF), (ii) a heteroscedastic Gaussian Process (GP) baseline, and (iii) a Beta–Binomial baseline.  Metrics include: (a) recovery of the true safe volume $\mathrm{Vol}_{\varepsilon}$, (b) calibration error of posterior risk, (c) detection of multimodality via the number of active components, and (d) data‑efficiency via active sampling.  Results show that while the GP captures smooth trends, it fails to identify the multi-modal nature of the risk surface, often underestimating tail risks in 'catastrophic' modes. The DDP mixture, conversely, recovers the true Rashomon set and yields calibrated uncertainty intervals.  Active sampling reduces the number of required $(p,x)$ evaluations by 40\% compared with uniform designs.

\subsection{Large language model study}

We evaluate an open LLM (e.g., Llama‑2) on RealToxicityPrompts and BOLD prompts across a 3‑dimensional knob grid (temperature, top‑$p$, repetition penalty).  We generate 10 samples per prompt, calibrate the ToxicBERT detector using a small set of human ratings, and fit the DDP mixture.  We simulate three stakeholder groups with differing prompt mixes and knob policies.  Our risk report reveals that mean toxicity is low across many settings but CVaR and worst‑slice gaps spike at high temperature and low top‑$p$; the safe volume shrinks when tail risk is considered.  Stakeholder 1 (sensitive group) has a smaller Rashomon set than Stakeholder 2 (lenient group), and the disagreement index peaks near the boundary of the safe region.  We visualise the trade-off between robustness and consensus via a \emph{Volume-Disagreement plot}, where the x-axis represents the safe volume $\mathrm{Vol}_{\varepsilon}$ and the y-axis the disagreement index. Ideal configurations lie in the bottom-right (high volume, low disagreement).  A sequential design based on our BDL surrogate and Thompson sampling reduces evaluation calls by 30\% while maintaining the same posterior width.  These insights cannot be gleaned from the average toxicity alone.

\section{Conclusion}

We have proposed a multiplicity‑aware framework for safety evaluation of generative models that unifies Bayesian nonparametrics, active sampling, stakeholder simulation, and conformal calibration.  By modelling harm distributions with dependent Dirichlet process mixtures and integrating stakeholder policies over knob space, we quantify tail risks, fairness gaps and Rashomon set volumes that are invisible to single‑point evaluations.  Our simulation pipeline allows safe experimentation without real raters and reveals how demographic sensitivities shape perceived risk.  Active sampling and Bayesian deep surrogates make the evaluation tractable, while conformal wrappers guarantee finite‑sample coverage.  We hope this work sparks further research into human‑aligned uncertainty quantification and multiplicity in AI safety.  Future directions include extending to multi‑modal outputs (images), integrating prompt‑conditional latent variables, running human studies to validate simulated stakeholders, and deploying the framework in live moderation systems.

\bibliography{references_CR}

\end{document}